%% file: main.tex
\documentclass[10pt,twocolumn,a4paper]{article}

\usepackage[a4paper,left=21mm,right=21mm,top=23mm,bottom=21mm,
  headheight=20pt,headsep=12pt,footskip=24pt,columnsep=6.2mm]{geometry}
\usepackage{fontspec}
\usepackage{amsmath}
\usepackage{unicode-math}
\usepackage{booktabs,graphicx,array,tabularx}
\usepackage{xcolor}
\definecolor{Brand}{HTML}{2859B8}
\definecolor{Ink}{HTML}{172131}
\definecolor{Muted}{HTML}{69717E}
\definecolor{Rule}{HTML}{CDD4DE}
\usepackage{titlesec,needspace,placeins}
\usepackage{caption}
\usepackage{fancyhdr}
\usepackage[numbers]{natbib}
\usepackage[hidelinks,unicode]{hyperref}
\hypersetup{pdftitle={Decoupled Predictive Decision Making: A Shared Conditional Predictor for Agents},
  pdfauthor={Lang Mo; Xingxuan Zhang; Qi Wang; Mingchao Hao; Hao Yuan; Chunjing Gan; Wang Lu; Peng Cui},
  pdfsubject={Technical report on agent decision making},pdfcreator={XeLaTeX}}
\graphicspath{{./}}

\newcommand{\brandwordmark}[1]{\includegraphics[width=#1,trim=9.5bp 4bp 2bp 1bp,clip]{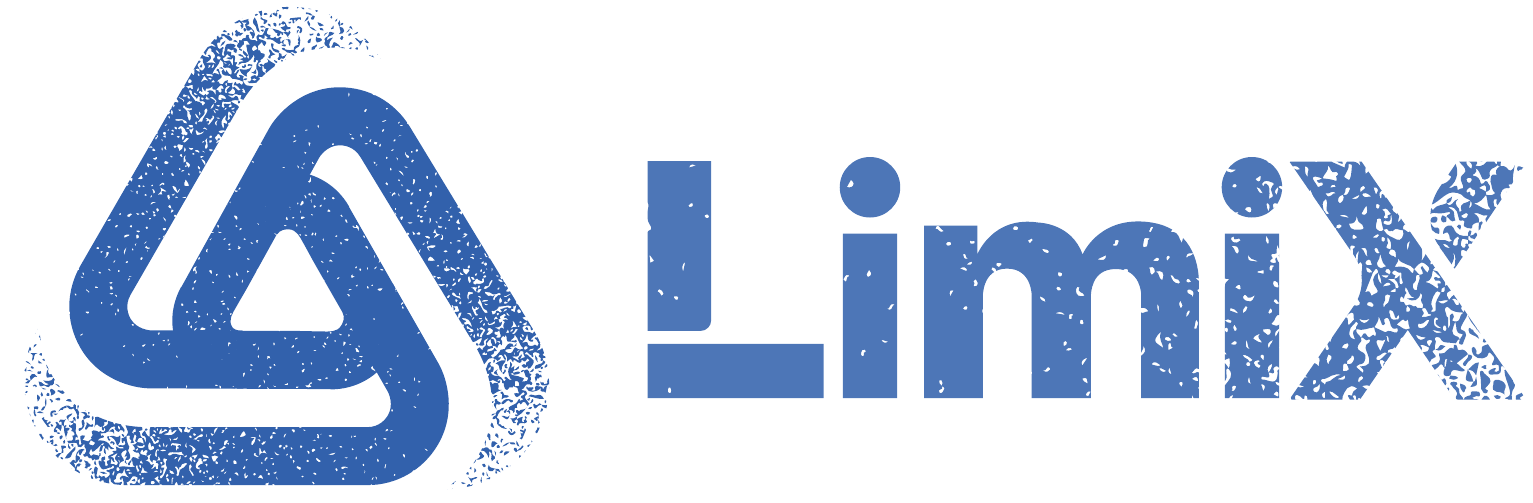}}
\newcommand{\reportpage}{\ifnum\value{page}<10 0\fi\thepage}
\fancypagestyle{opening}{\fancyhf{}
  \fancyfoot[R]{{\sffamily\fontsize{8}{9}\selectfont\color{Ink}\reportpage}}
  }
\fancypagestyle{plain}{\fancyhf{}
  \fancyfoot[R]{{\sffamily\fontsize{8}{9}\selectfont\color{Ink}\reportpage}}
  }

\titleformat{\section}
  {\sffamily\fontsize{11.1}{13.3}\selectfont\bfseries\color{Ink}\raggedright}
  {\textcolor{Brand}{\thesection}}{0.65em}{}
\titleformat{\subsection}
  {\sffamily\fontsize{10.1}{12.3}\selectfont\bfseries\color{Ink}\raggedright}
  {\textcolor{Brand}{\thesubsection}}{0.55em}{}
\titleformat{\subsubsection}
  {\sffamily\fontsize{9.7}{11.8}\selectfont\bfseries\color{Ink}\raggedright}
  {\thesubsubsection}{0.55em}{}
\titlespacing*{\section}{0pt}{13pt plus 2pt minus 1pt}{5pt}
\titlespacing*{\subsection}{0pt}{10pt plus 2pt minus 1pt}{4pt}
\titlespacing*{\subsubsection}{0pt}{8pt plus 1pt}{3pt}
\titlespacing*{\paragraph}{0pt}{5pt plus 1pt}{0.5em}
\begin{document}
\fontsize{10}{12.9}\selectfont
\raggedbottom
\twocolumn[{\begin{minipage}{\textwidth}
  \noindent\brandwordmark{29mm}\par
  \vspace{9pt}
  {\color{Brand}\hrule height 0.6pt}
  \vspace{19pt}
  \begin{center}
  {\fontsize{21}{25.5}\selectfont\bfseries\color{Ink}
   Decoupled Predictive Decision Making:\\
   A Shared Conditional Predictor for Agents\par}
  \vspace{12pt}
  {\fontsize{11}{15}\selectfont\color{Ink}
   Lang Mo\textsuperscript{1,*}\hspace{1.2em}Xingxuan Zhang\textsuperscript{1,*,\dagger}\hspace{1.2em}Qi Wang\textsuperscript{1}\hspace{1.2em}Mingchao Hao\textsuperscript{1}\par
   \vspace{1pt}
   Hao Yuan\textsuperscript{1}\hspace{1.2em}Chunjing Gan\textsuperscript{1}\hspace{1.2em}Wang Lu\textsuperscript{1}\hspace{1.2em}Peng Cui\textsuperscript{2,\dagger}\par}
  \vspace{5pt}
  {\fontsize{9.5}{12}\selectfont\color{Ink}
   \textsuperscript{1}Stable AI\hspace{1.8em}\textsuperscript{2}Tsinghua University\par}
  \vspace{3pt}
  {\fontsize{8.5}{11}\selectfont\color{Muted}
   \textsuperscript{*}Equal contribution.\hspace{1.4em}\textsuperscript{\dagger}Corresponding authors.\par
   \vspace{2pt}
   Contact: \href{mailto:12332894@mail.sustech.edu.cn}{12332894@mail.sustech.edu.cn}\hspace{1.2em}\href{mailto:xingxuanzhang@hotmail.com}{xingxuanzhang@hotmail.com}\hspace{1.2em}\href{mailto:cuip@tsinghua.edu.cn}{cuip@tsinghua.edu.cn}\par}
  \end{center}
  \vspace{9pt}
  \begin{minipage}{\textwidth}
  {\sffamily\fontsize{8.2}{10}\selectfont\bfseries\color{Brand}ABSTRACT}\par
  \vspace{4.5pt}
  {\fontsize{9.6}{12.1}\selectfont\setlength{\parindent}{0pt}
   \input{sections/abstract_v1.3}\par}
  \end{minipage}
  \vspace{13pt}
  {\color{Rule}\hrule height 0.35pt}
  \vspace{14pt}
\end{minipage}
}]
\thispagestyle{opening}
\input{body}
\end{document}

%% file: sections/abstract_v1.3.tex
Agent decisions depend on empirical relationships that deployment
observations reveal and task descriptions alone may leave unresolved.
Learning these relationships requires a mechanism that turns historical
evidence into predictions. We formulate agent decision making through
general-purpose conditional inference, introducing \emph{Decoupled Predictive
Decision Making} (DPDM), an architecture in which a shared pretrained
learner serves heterogeneous decision roles. Language models provide
semantic representations of current and historical records, while a general-purpose conditional predictor predicts task-defined targets
from the represented evidence. Pretraining supplies a reusable inference
capability; deployment evidence instantiates problem-specific predictions
without updating either model's neural parameters. Tool judgments,
candidate preferences, and reasoning assessments thus share a common
evidence-and-query interface, with predictions mapped to choices by a
decision rule. This architecture establishes complementary sources of
agent capability: semantic knowledge acquired through language pretraining
and empirical inference acquired through structured-data pretraining.
We instantiate the conditional predictor with LimiX-2, pretrained on structured data,
which is shared across all evaluated decision roles. Across five task categories,
DPDM accepts
representations from semantic encoders such as Qwen3.5.
With Qwen3.5-27B as the semantic encoder, DPDM raises mean accuracy
across 50 benchmarks from 78.80\% with few-shot prompting to 84.92\%
under their respective evaluated history budgets, and reaches 82.33\% with identical
histories and queries. On ToolPRMBench's BFCL and ToolSandbox tasks,
absolute accuracy gains are 34.26\% and 28.87\%, respectively,
at the methods' respective evaluated history budgets. Accuracy improves as history grows.
With identical imbalanced histories, the absolute gain in mean balanced accuracy is
14.51\% across five reasoning-step assessment tasks and
four encoders.
DPDM makes learning from experience an independently improvable
foundation-model capability, creating a shared route for advances in
conditional prediction to improve decisions across agent tasks.

%% file: body.tex
\input{tables/review_numbers}
\input{tables/completion_numbers}
\input{sections/introduction_v1.3}
\input{sections/method_v1.3}
\FloatBarrier
\input{sections/experiments_current}

\FloatBarrier
\input{sections/discussion_conclusion}
\FloatBarrier
\bibliographystyle{plainnat}
\begingroup\footnotesize
\setlength{\bibsep}{1.5pt}
\bibliography{method_references,experiment_references,v1.3_references,agent_references}
\endgroup
\clearpage
\appendix
\fontsize{10}{12.5}\selectfont
\input{sections/method_appendix}
\input{sections/experiments_appendix}

\input{sections/decision_quality}
\input{sections/decision_theory}
\FloatBarrier

%% file: tables/review_numbers.tex
\newcommand{\QwenZero}{66.98}
\newcommand{\QwenHistory}{81.75}
\newcommand{\QwenSD}{0.02}
\newcommand{\QwenWins}{51}
\newcommand{\QwenTies}{0}
\newcommand{\QwenLosses}{1}
\newcommand{\FrequencyAccuracy}{50.52}
\newcommand{\ToolFew}{84.00}
\newcommand{\ToolHistory}{92.31}
\newcommand{\CodeFew}{80.76}
\newcommand{\CodeHistory}{85.61}
\newcommand{\FactFew}{69.13}
\newcommand{\FactHistory}{77.62}
\newcommand{\SafetyFew}{85.36}
\newcommand{\SafetyHistory}{87.34}
\newcommand{\LanguageFew}{87.68}
\newcommand{\LanguageHistory}{89.37}

%% file: tables/completion_numbers.tex
\newcommand{\JevZeroAccuracy}{75.94}
\newcommand{\HistoryUseDatasetCount}{54}
\newcommand{\HistoryUseTaskCount}{89}

%% file: sections/introduction_v1.3.tex
\section{Introduction}
\label{sec:introduction}

\textbf{A generalist agent must learn how to decide, as well as understand
what it is asked to do.} Large language models support few-shot
adaptation~\cite{brown2020fewshot}, multi-step reasoning~\cite{wei2022cot},
and interactive tool use~\cite{yao2023react}. The value of a particular
choice depends on the conditions under which it is made. Adaptive retrieval
studies show that queries benefit from different retrieval
strategies~\cite{jeong2024adaptive}, while work on test-time computation
shows that effective allocation depends on problem difficulty~\cite{snell2025scaling}.
These findings motivate a distinction between representing a decision
problem and learning the relationships that determine an appropriate
choice. Semantic competence expands the range of problems an agent can
interpret, while learning from feedback determines how effectively it can adapt
its decisions to those problems.

\textbf{Effective learning from experience connects historical feedback
to current decisions.} Retrieved trajectories and reflective feedback can
improve subsequent behavior~\cite{zheng2024synapse,shinn2023reflexion},
and language models can learn numerical relationships from in-context
examples~\cite{vacareanu2024regressor}. The resulting decision
depends on how the learner interprets the evidence. A previously successful
action may be inappropriate under different conditions, while a frequently
selected action may appear effective because of the situations in which it
was used. The agent must identify conditional relationships across
observations and apply them to the current decision. We distinguish two sources of decision error: information that the available
history does not reveal, and information that the learning mechanism fails to use. Growing histories tests the value
of additional evidence (Section~\ref{sec:exp_history}), while comparing
learners on identical histories tests how effectively they use that evidence
(Section~\ref{sec:exp_generality}). 
This distinction separates acquiring more informative experience from
improving the computation that converts existing evidence into a choice.

Existing approaches organize this computation in different ways.
ExpeL extracts insights and recalls experience~\cite{zhao2023expel};
Rememberer updates an external experience memory using
reinforcement-learning feedback~\cite{zhang2023rememberer}.
Algorithm Distillation learns from reinforcement-learning
histories~\cite{laskin2022ad}, DPT predicts optimal actions from interaction
contexts~\cite{lee2023dpt}, and PreDeToR pretrains reward prediction across
related decision tasks~\cite{mukherjee2024predetor}. SayCan combines language
planning with learned skill values~\cite{ahn2022saycan}.
These precedents establish several forms of experience-based adaptation
and computational specialization. Within language-model-centered systems, reasoning and action generation
remain closely integrated~\cite{yao2023react}, and retrieved experience can
be interpreted through the same language model~\cite{zhao2023expel}.
This organization couples semantic interpretation, empirical inference,
and answer generation within one pretrained mechanism. The mechanism for learning
deployment-specific relationships is not exposed as a separately
replaceable component, limiting the ability to improve empirical inference
independently of the language policy. Semantic competence alone does not
establish how effectively that mechanism uses a given history. The architectural choice is whether the language model should both
represent a decision problem and learn its deployment-specific relationships,
or whether a shared conditional learner should infer those relationships
from historical feedback. We develop an explicit separation between
semantic representation and history-conditioned prediction, making each a
replaceable and independently improvable component of the decision
architecture.

\needspace{4\baselineskip}
\textbf{We introduce Decoupled Predictive Decision Making (DPDM), a framework that reformulates
task-defined agent decisions as conditional prediction problems.}
Each decision specifies its visible context, candidate descriptions, and
target, while historical records provide observations of that target under
related conditions. A language model serves as the semantic encoder,
representing the problem and candidate answers. A shared conditional
predictor uses the represented history to estimate the unknown target
for the current decision, and a decision rule converts the prediction into
a choice. Pretraining a conditional predictor across datasets provides a foundation
for this construction, as established by Prior-Data Fitted Networks and
Contextual Mechanism Networks~\cite{hollmann2025tabpfn,limix2026}. DPDM connects
this learning capability to semantically rich agent decisions through
representations of past experience and the current decision context.
Tool applicability, candidate
preference, and reasoning assessment become different prediction targets
for the same learning mechanism. When action--outcome observations are
available, the conditional predictor can estimate consequences for
objective-dependent selection.

We instantiate the conditional predictor with LimiX-2~\cite{zhang2025limix,limix2026},
a structured-data foundation model pretrained for conditional prediction.
It conditions on the available historical feedback to estimate the
current target. Both neural models remain frozen as the history changes.
The architecture exposes
separate routes to improvement: better semantic representations can reveal
relevant distinctions, and advances in conditional prediction can improve
target estimates using those distinctions and historical
feedback. Predictions can also support different objectives when they
retain the information required by the decision rule.

Across decision categories and semantic encoders, a single frozen
conditional predictor improves mean decision accuracy over few-shot
prompting. Gains persist with identical histories and queries, showing
that additional records alone do not explain the advantage. On
ToolPRMBench BFCL, DPDM achieves an absolute accuracy gain of
34.26\% over few-shot prompting at the methods' respective evaluated history budgets. Growing-history and imbalanced-history
evaluations show benefits from both accumulating experience and making
better use of the same feedback. Controls demonstrate complementary
contributions from historical input--target relationships, distinct
observations, and semantic features. In a separate timing study with
matched histories and query sets, the equal-task mean reduction in
measured full-pipeline GPU time, including historical preparation,
is 80.3\% relative to cached few-shot prompting. These findings support
an architecture in which the mechanism for learning from experience is
shared across decision roles, while each task supplies its own evidence
and query without task-specific neural-parameter updates.

\textbf{Learning from experience becomes an explicit target of
foundation-model development.} Systems such as Voyager retain accumulated
experience as reusable skills~\cite{wang2023voyager}. DPDM
develops a complementary route: a shared conditional predictor turns
task-relevant observations into new judgments. Improving that predictor
can benefit multiple decision roles through the same interface, alongside
advances in the semantic encoder that represents the problem.

%% file: sections/method_v1.3.tex
\input{sections/framework_figure}
\section{Method}
\label{sec:method}

Our architecture separates understanding a decision problem from using
historical feedback to resolve it. A semantic encoder represents the
current context and candidate choices, a shared conditional predictor
estimates the decision target from represented historical examples, and
a decision rule selects the answer (Figure~\ref{fig:framework}).
We instantiate the conditional predictor with LimiX-2. Every task uses the
same checkpoint with its own history and current query.

\subsection{Decision Making as History-Conditioned Prediction}
\label{sec:capability}

A task specification $\mathcal T$ defines the observable input fields,
the meaning of the prediction target, and the candidate answer set
$\mathcal Y_{\mathcal T}(r)$ for a record $r$. The record contains information
available before the target is known, including the task context and,
when applicable, the visible interaction trajectory and ordered candidate
descriptions. The specification is provided by the task and
applied consistently to historical and query records.

Let $S=\{(r_i,y_i)\}_{i=1}^{M}$ denote the history of $M$ records and their
observed targets. Historical feedback is recorded as the target $y_i$
for each record $r_i$.
A query record $r$ describes the current decision
whose target $y$ is unknown. In the primary setting,
$y$ is a task-defined answer, correctness judgment, or candidate preference.
For candidate selection, historical labels identify the candidate that
satisfies the task criterion. A trajectory contained in $r$ describes the
current decision context; the history $S$ supplies completed examples
from which the relevant input--target relationship is inferred.

The prediction problem is to estimate $q_S(y\mid r,\mathcal T)$.
Different decision roles define different records and targets, while
retaining the same inputs: represented historical examples, their
observed targets, and a query. Historical and query records share the same
target semantics within each task, and the conditional predictor returns
probabilities over that task's answer set.

\subsection{A Shared Conditional Predictor}
\label{sec:pretrained}
\label{sec:conditional_learning}

The conditional predictor $F_\theta$ takes three inputs: represented historical records,
their observed targets, and a represented query. It returns probabilities
for the query's target. A semantic encoder and a feature adapter prepare
these inputs, allowing the same conditional predictor to accept records
from different tasks and semantic encoders.

\paragraph{Representing historical and query records.}
\label{sec:representation}
At each decision, $r$ and $\mathcal T$ jointly specify the observable
decision state: the task description, current observations, and, when
applicable, descriptions of the available actions or other candidates.
A frozen semantic encoder $J_\phi$ represents each record in the context
of its target question and candidate descriptions. This interface places decision-specific structure in the represented
inputs and historical evidence, allowing the shared conditional predictor
to infer the answer without task-specific updates to its neural parameters.
Let $b_{\mathcal T}(r)$ collect available native fields and lexical information. A feature
adapter maps these inputs into a structured representation:
\begin{equation}
 z_S(r)=A_{\psi_S}\!\left(J_\phi(r;\mathcal T),b_{\mathcal T}(r)\right).
 \label{eq:features}
\end{equation}
Here $\psi_S$ denotes any state fitted from historical inputs, such as a
feature transformation. The same fitted mapping is applied to historical and
query records. Historical targets are excluded from semantic record
encoding and supplied separately as supervision; query targets remain
unavailable throughout representation fitting and prediction.

\paragraph{Inferring a target from historical examples.}
The conditional predictor receives the represented history and query:
\begin{equation}
 \widetilde S=\{(z_S(r_i),y_i)\}_{i=1}^{M},
 \label{eq:represented_support}
\end{equation}
\begin{equation}
 q_S(\cdot\mid r,\mathcal T)=F_\theta\!\left(\widetilde S,z_S(r)\right).
 \label{eq:prediction}
\end{equation}
Equation~\ref{eq:prediction} returns probabilities over the task's candidate
answers. The task specification determines how records are represented
and how outputs map to answers. Appendix~\ref{app:implementation} gives
the decoding and normalization details.

\paragraph{Pretrained instantiation.}
We use LimiX-2 as $F_\theta$, pretrained for
conditional prediction over structured data~\cite{limix2026}, independently
of the evaluated decision tasks. Both the semantic
encoder parameters $\phi$ and the conditional predictor parameters $\theta$
remain fixed. Alternative pretrained conditional predictors can fill this role
when they support the required representations, target types,
and output distributions. Feature construction, output coding, and model
configuration are specified in Appendix~\ref{app:implementation}.

\subsection{Adapting to New Observations}
\label{sec:update}

Deployment adaptation changes the history supplied to the conditional predictor.
Newly obtained labels or observed outcomes provide additional
records $\Delta S$, yielding
\begin{equation}
 S^+=S\cup\Delta S,\qquad \phi^+=\phi,\qquad\theta^+=\theta.
 \label{eq:evidence_update}
\end{equation}
The history-dependent adapter state $\psi_{S^+}$ and the resulting
prediction $q_{S^+}$ may change. Any required input transformations are
refitted on the updated historical inputs and applied consistently to
historical and query records. Semantic representations can be reused when the
encoder and record specification are unchanged.

Section~\ref{sec:exp_history} evaluates this adaptation by expanding
historical sets and predicting the same queries.

%% file: sections/framework_figure.tex
\begin{figure*}[t]
\centering
\includegraphics[width=0.98\textwidth]{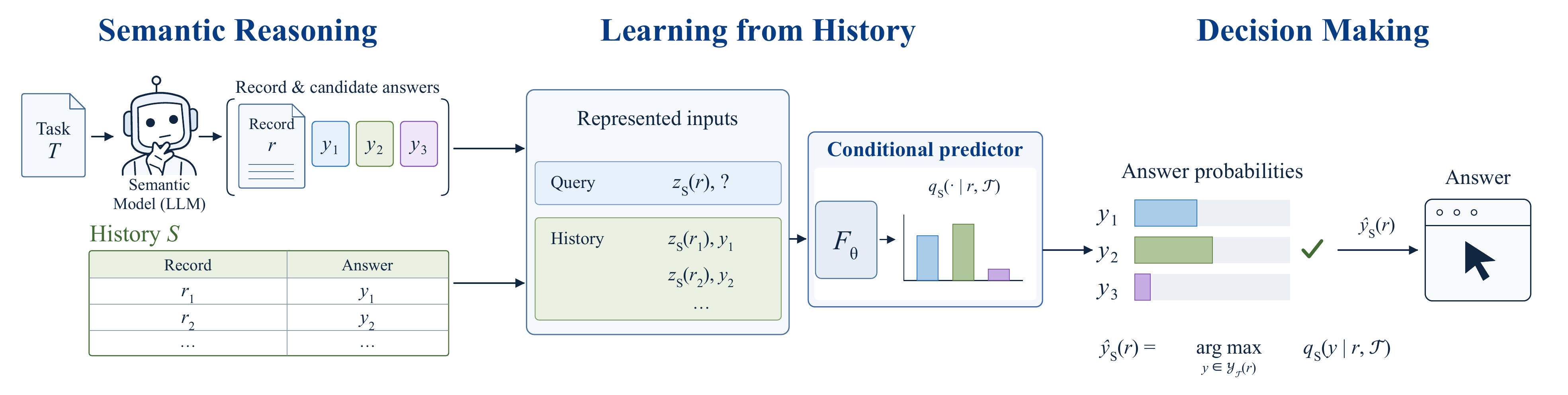}
\caption{DPDM uses historical feedback to make the current decision.
A semantic encoder represents each record, and a shared conditional predictor,
instantiated here with LimiX-2, estimates the current target from history.
The diagram follows the main benchmark:
historical targets are supplied separately from record representations,
and the highest-probability candidate answer is selected.}
\label{fig:framework}
\end{figure*}

%% file: sections/experiments_current.tex
\section{Experiments}
\label{sec:experiments}

We test whether a shared frozen predictor supports heterogeneous
decision roles and semantic encoders, whether additional history improves
decisions without neural-parameter updates, and which properties of the
evidence account for those gains. Across 50 benchmarks, we compare DPDM
with direct prompting of the same Qwen3.5 and Llama-3.1 models, including
identical-history comparisons. Imbalanced histories and controlled
interventions examine the roles of historical relationships, distinct
observations, and semantic representations.

\begin{table}[t]
\centering
\caption{Decision accuracy (\%) across 50 benchmarks.
Each language model is evaluated through direct prompting and as a
semantic encoder in DPDM.
Benchmark scores average task configurations; the overall score
weights benchmarks equally, and category scores use the same rule within
each category. DPDM results average three sampled-history runs;
$\pm$ denotes their sample standard deviation.}
\label{tab:main_results}
\small
\setlength{\tabcolsep}{3pt}
\input{tables/qwen_comparison}
\end{table}

\subsection{Tasks, Models, and Evaluation}
\label{sec:exp_setup}

Our evaluation covers 50 benchmarks, which comprise 82 task configurations across five categories:
agent tasks, knowledge QA, evidence consistency, safety and reliability,
and language understanding. Each task defines
a decision target for the current case. We evaluate whether an agent selects the task-appropriate answer,
candidate, or judgment. DPDM reformulates each such decision as a
conditional prediction problem: the current case defines the query,
historical records supply observed decision targets, and the shared
predictor estimates the target before a decision rule selects the answer.
Decision accuracy measures agreement with the task's reference target.

The agent-related tasks include intent recognition, tool applicability,
response-mode judgments, intermediate-step evaluation, and response
preference. For example, BFCL Callability asks whether any supplied
function is applicable to a user request~\cite{patil2025bfcl}, while a
response-mode task derived from ToolACE distinguishes text, a single
tool call, and multiple calls~\cite{liu2024toolace}.
Appendix~\ref{app:inventory} summarizes the benchmarks,
decision targets, and historical feedback used in the evaluation.

We use Qwen3.5-4B, 9B, and 27B and
Llama-3.1-8B-Instruct~\cite{qwen35model,meta2024llama31}.
Each model is evaluated through direct prompting and as the semantic
encoder in DPDM. Zero-shot prompting supplies only the current
input. Few-shot prompting adds historical inputs and their observed targets.

DPDM assigns semantic interpretation and empirical inference to
different pretrained capabilities. Few-shot prompting uses the language
model to interpret the history and produce the decision; DPDM uses it
to represent the decision problem and assigns history-conditioned
prediction to a shared learner. In DPDM, the language model encodes each input and its
candidate descriptions independently. Historical targets are supplied
separately to the shared conditional predictor, not included in the
semantic-encoding prompts. The predictor uses the representations and
historical targets to predict the current target. We select the answer
with the highest predicted probability.
Each encoder is evaluated separately with the same LimiX-2 checkpoint,
and both neural models' weights remain frozen throughout the experiments.
We denote an instantiation by DPDM\_Encoder, for example
DPDM\_Qwen3.5-27B. Where a table row or figure panel already identifies
the encoder, the abbreviated label DPDM denotes that instantiation.
Where used in figure legends, ``+LimiX-2'' denotes the same
encoder--predictor combination.

Jev API v1.13.0 provides an external decision-service
reference~\cite{typesafe2026jev}, evaluated in zero-shot and few-shot
settings. Appendix~\ref{app:exp_protocol} gives the readout settings and
service histories. Supplementary model comparisons appear in
Appendix~\ref{app:extended_results}.

History size $M$ is the total number of historical examples. All methods
answer the same test queries within each task, and these queries are
kept separate from the history. The main comparison uses each language
model's largest completed few-shot history under the context, memory,
and runtime budgets of the evaluation. For DPDM, we use the
largest history with complete results for every evaluated encoder under the same task-specific history budget.
Processing a historical record has different costs in the two
architectures, so their completed histories can contain different numbers
of records.

We average task configurations within each benchmark and then give
each benchmark equal weight. 
Appendix~\ref{app:exp_protocol} gives the sampling and history-budget
details.

\subsection{A Shared Conditional Predictor Across Encoders and Tasks}
\label{sec:exp_generality}

\begin{figure}[t]
\centering
\includegraphics[width=\columnwidth]{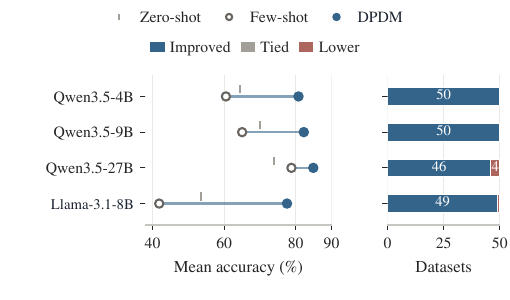}
\caption{DPDM improves accuracy across Qwen and Llama encoders.
Left: mean accuracy over 50 benchmarks; connected points compare few-shot
prompting with DPDM using the same encoder, and short marks show zero-shot accuracy.
Right: numbers of datasets with higher, equal, or lower accuracy than
few-shot prompting.}
\label{fig:model_comparison}
\end{figure}

\begin{figure*}[t]
\centering
\includegraphics[width=\textwidth]{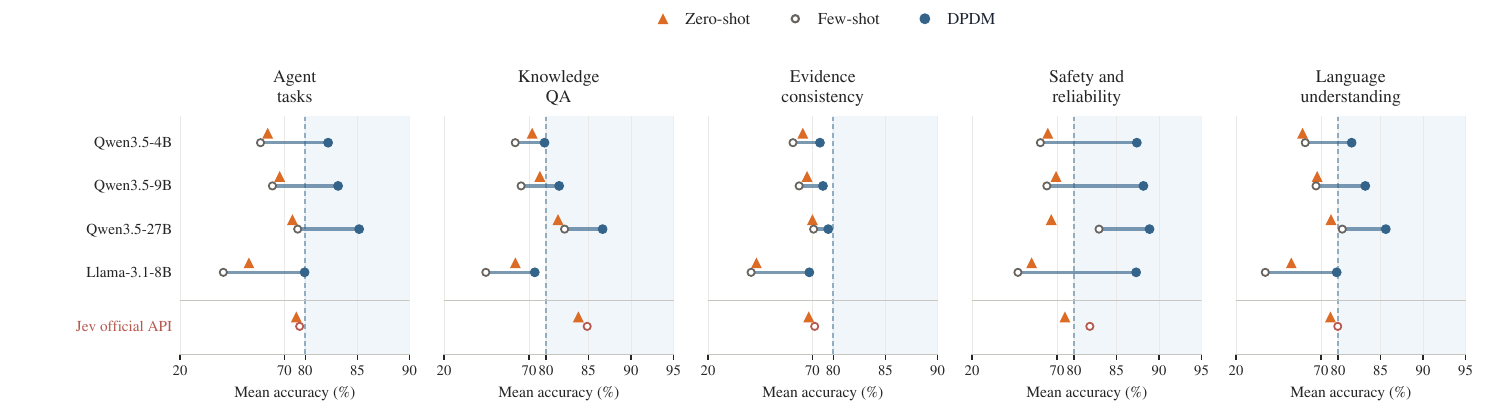}
\caption{Decision accuracy across five task categories.
Each panel shows one category. Open and filled circles compare few-shot
prompting with DPDM using the same encoder; triangles show zero-shot accuracy.
The bottom row gives the Jev API reference. Within each panel, the shaded
region above 80\% uses five times the horizontal scale; no intervals are omitted.}
\label{fig:decision_types}
\end{figure*}

This comparison tests reuse of one predictive core across semantic
encoders and decision roles. DPDM exceeds few-shot prompting for all four
encoders in all five categories at the reported endpoint budgets. Its
overall gains persist on identical histories, showing that additional
records alone do not explain the improvement.

With Qwen3.5-27B as the semantic encoder, DPDM reaches 84.92\% mean accuracy, compared with 78.80\% for few-shot prompting
(Table~\ref{tab:main_results} and Figure~\ref{fig:model_comparison}).
The corresponding DPDM and few-shot accuracies are 80.78\% versus 60.51\% for Qwen3.5-4B and
82.27\% versus 65.03\% for Qwen3.5-9B. DPDM improves accuracy
on 46 of the 50 datasets with the 27B encoder and on all 50 with each
of the two smaller Qwen encoders. With Llama-3.1-8B, accuracy rises
from 41.87\% with few-shot prompting to 77.57\% with DPDM,
improving on 49 of the 50 datasets. The external Jev API reference
reaches \JevZeroAccuracy\% in the zero-shot setting and 79.30\% in
the few-shot setting.

Few-shot prompting is less accurate than zero-shot prompting for
Qwen3.5-4B, Qwen3.5-9B, and Llama-3.1-8B in this evaluation.
DPDM exceeds both prompting results for all four encoders.

Encoding historical records individually allows DPDM to use
histories beyond a single language-model prompt. On FaithDial,
DPDM\_Qwen3.5-27B reaches 83.53\% accuracy with 8,192 historical records,
compared with 54.70\% for Qwen3.5-27B few-shot prompting with 512 records.

We also compare DPDM with few-shot prompting using identical
historical records and queries to examine how effectively they use the same feedback. This
comparison covers all 50 benchmarks and 82 task configurations. For each
encoder--task pair, we use its largest completed common history, keeping
the examples, their order, and their targets identical.
DPDM reaches 82.33\%
with Qwen3.5-27B, compared with 78.80\% for few-shot prompting.
The paired scores are 78.37\% versus 60.51\% for Qwen3.5-4B,
80.08\% versus 65.03\% for Qwen3.5-9B, and 73.83\% versus 41.87\%
for Llama-3.1-8B.

Figure~\ref{fig:decision_types} reports the main comparison across the
five task categories. For every encoder, DPDM has
higher mean accuracy than direct few-shot prompting in each category.
In agent tasks,
Qwen3.5-9B improves from 64.30\% to 83.16\%, Qwen3.5-27B from 76.41\%
to 85.18\%, and Llama-3.1-8B from 40.77\% to 79.73\%.
For Qwen3.5-27B, tool-use assessment on ToolPRMBench BFCL improves from 42.59\% with few-shot prompting to 76.85\% with DPDM.
CodeCritic Code accuracy rises from 76.82\% to 90.85\% with the same encoder.
Additional language-model and DPDM comparisons, including four
StartLux encoder sizes, are reported in Appendix~\ref{app:extended_results}.

\subsection{Learning from Growing Histories}
\label{sec:exp_history}

\begin{figure}[!htbp]
\centering
\includegraphics[width=0.80\columnwidth]{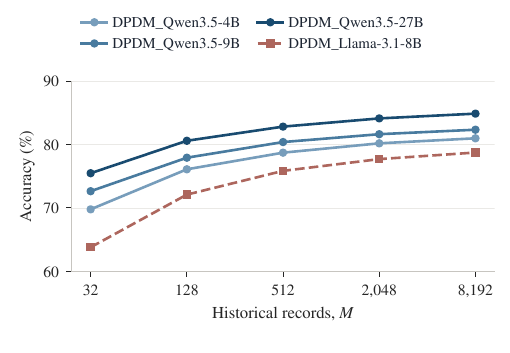}
\caption{DPDM makes more accurate decisions as history grows.
Each curve pairs LimiX-2 with the frozen semantic encoder named in the
legend, averaging accuracy over the same 21 datasets.
The horizontal axis counts records on a logarithmic scale.}
\label{fig:qwen_history}
\end{figure}

This experiment tests whether additional observations improve DPDM
without changing either neural model. On the common benchmark cohort,
mean accuracy increases at every evaluated history size for all four
encoders, supporting adaptation through the supplied evidence. We vary
the number of historical examples while keeping the test queries and
both neural models' weights fixed. Figure~\ref{fig:qwen_history}
reports accuracy on the same 21 benchmarks and 29 task configurations
at 32, 128, 512, 2,048, and 8,192 historical records. This analysis
includes tasks with results at all five history sizes.
Appendix~\ref{app:task_curves} lists the tasks and describes the sampling.

\begin{figure*}[t]
\centering
\includegraphics[width=\textwidth]{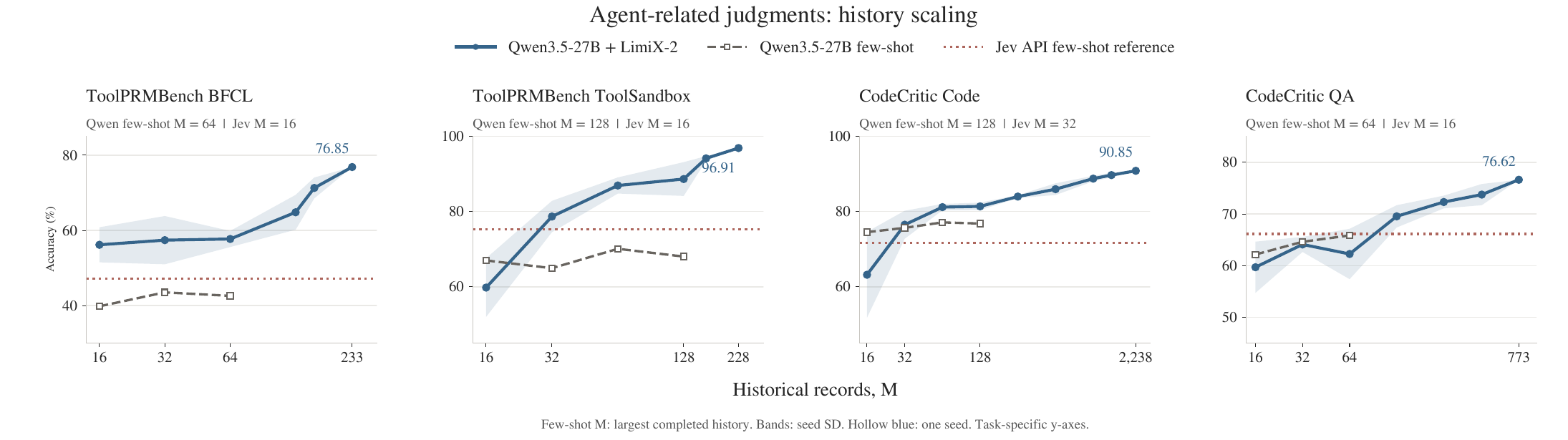}
\caption{Additional history improves four agent-related judgments.
DPDM uses Qwen3.5-27B representations; bands show standard deviations
across three history samples. Dashed curves show measured Qwen few-shot
accuracy; dotted lines show Jev few-shot references on the same queries.
The displayed histories start at 16 records. History sizes and reference
budgets are listed in Appendix~\ref{app:task_curves}.}
\label{fig:task_scaling}
\end{figure*}

Accuracy increases from 69.78\% to 80.98\% with Qwen3.5-4B,
72.65\% to 82.33\% with Qwen3.5-9B, and 75.48\% to 84.87\% with
Qwen3.5-27B as history grows from 32 to 8,192 records. Llama-3.1-8B
increases from 63.81\% to 78.74\%. Accuracy rises at each history size
for all four encoders.

We examine this history-growth pattern in four agent-related tasks
using DPDM\_Qwen3.5-27B (Figure~\ref{fig:task_scaling}).
ToolPRMBench's BFCL and ToolSandbox
suites assess tool-use steps~\cite{li2026toolprmbench}; CodeCritic
contributes its Code and QA assessment tasks~\cite{zhang2025codecritic}.

At the largest evaluated histories, DPDM\_Qwen3.5-27B reaches 76.85\%
on BFCL versus 42.59\% for Qwen few-shot prompting, and 96.91\% on
ToolSandbox versus 68.04\%. The paired scores are 90.85\% versus 76.82\%
on CodeCritic Code and 76.62\% versus 65.85\% on CodeCritic QA.
The dashed curves show Qwen few-shot accuracy at the measured history
sizes; horizontal dotted lines show Jev few-shot accuracy on the same
queries using task-specific prompt histories. Their history
sizes are given in Appendix~\ref{app:task_curves}.

\input{sections/agent_analysis_current}
\input{sections/qwen_controls}
\FloatBarrier
\input{sections/cost_current}

%% file: tables/qwen_comparison.tex
\begin{tabular}{@{}lrrr@{}}
\toprule
Encoder & Zero-shot & Few-shot & DPDM \\
\midrule
Qwen3.5-4B & 64.56 & 60.51 & \textbf{80.78} $\pm$ 0.11 \\
Qwen3.5-9B & 69.92 & 65.03 & \textbf{82.27} $\pm$ 0.15 \\
Qwen3.5-27B & 74.02 & 78.80 & \textbf{84.92} $\pm$ 0.14 \\
Llama-3.1-8B & 53.49 & 41.87 & \textbf{77.57} $\pm$ 0.13 \\
Jev API & 75.94 & 79.30 & --- \\
\bottomrule
\end{tabular}

%% file: sections/agent_analysis_current.tex
\subsection{Making Decisions under Imbalanced Histories}
\label{sec:exp_evidence}

When most historical examples have the same answer, few-shot prompting
can favor that answer on new inputs that require a different judgment.
This tendency is known as \emph{majority-label bias}, and prior work shows
that the frequency and order of in-context labels affect a language
model's predictions~\cite{zhao2021calibrate,fei2023labelbias}.
BIG-Bench Mistake provides a direct example: each query contains an
original question, preceding reasoning, and a current step whose
correctness needs to be judged~\cite{tyen2024mistake}.
We include all five tasks---Dyck languages, logical deduction, multistep
arithmetic, object tracking, and word sorting---and all 7,032 query steps.
For each of the four encoders and each task, few-shot prompting and DPDM receive
identical histories and queries at their largest completed common
history size. The histories contain 128--1,024 records,
of which 66.41--95.31\% are correct steps.

Correct steps also dominate the query sets, so always answering
\emph{correct} can yield high ordinary accuracy. We use
\emph{balanced accuracy} (BA) to give equal weight to correct and
incorrect steps~\cite{brodersen2010balanced}:
\[
\mathrm{BA}=\tfrac{1}{2}\left(
\mathrm{Acc}_{\mathrm{correct}}+\mathrm{Acc}_{\mathrm{incorrect}}
\right).
\]
The two terms are the fractions of correct steps accepted and incorrect
steps rejected. A constant answer scores 50\%, making the ability to
distinguish the two classes visible even when one class is rare.

\begin{figure}[!htbp]
\centering
\includegraphics[width=0.84\columnwidth]{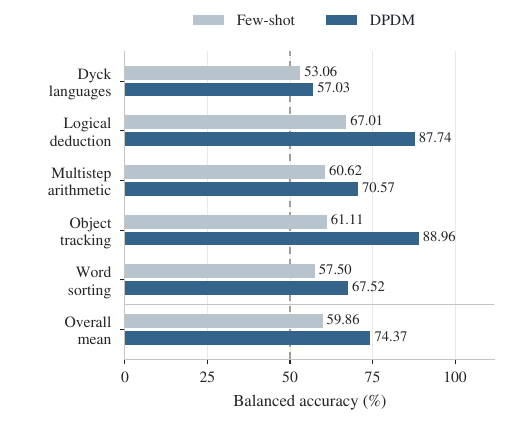}
\caption{Decisions under imbalanced histories in BIG-Bench Mistake.
Each task averages the three Qwen3.5 encoders and Llama-3.1-8B. The overall mean weights
the five tasks equally. The dashed line marks the 50\% BA of a constant answer.}
\label{fig:history_bias}
\end{figure}

Mean balanced accuracy rises from 59.86\% with few-shot prompting to
74.37\% with DPDM. The task-mean score improves on all five tasks
(Figure~\ref{fig:history_bias}), with gains in 18 of the 20
encoder--task pairs. Appendix~\ref{app:history_bias} reports every pair,
including ordinary accuracy and the two class-specific scores.

On object tracking, both methods receive the same 689 historical steps,
of which 77.36\% are correct, and answer the same 462 query steps.
Qwen3.5-27B few-shot prompting reaches 69.07\% BA and 84.85\% ordinary accuracy.
DPDM\_Qwen3.5-27B reaches 92.62\% BA and 94.37\% ordinary accuracy.
It correctly rejects 93 of the 104 incorrect steps, compared with 42 for
few-shot prompting, while correctly accepting 95.81\% of correct steps.

%% file: sections/qwen_controls.tex
\subsection{Roles of Evidence and Semantic Representation}
\label{sec:exp_controls}

DPDM organizes decision making around two complementary capabilities:
semantic representation of the decision context and conditional inference
from historical evidence. We test the empirical basis of this division
by intervening on the information supplied to the shared predictive core. Correct input--target pairings determine
which relationships the predictor can learn; distinct labeled observations
provide evidence beyond repeated rows; semantic features expose
distinctions relevant to predicting the target. The results support all
three contributions in the evaluated settings, explaining why useful
representations and informative history work together in the shared
prediction interface.

We next examine how historical feedback and semantic features contribute
to decision accuracy. All controls use Qwen3.5-27B as the semantic encoder and the same frozen
LimiX-2 checkpoint. We report task-averaged accuracy, with the mean and
sample standard deviation across three sampled histories.

We test how historical input--target pairings affect accuracy by
permuting historical targets across fixed inputs on six tasks:
BFCL Callability, ToolPRMBench
ToolSandbox, CodeCritic Code, PRM800K, FaithDial, and TREC coarse.
Each condition uses the same 128 historical inputs per task, a fixed
set of 737 queries across the six tasks, and the same fitted
representation. Permuting historical labels breaks input--target
pairings while preserving the number of examples in each class.

\begin{table}[!htbp]
\centering
\caption{Correct historical input--target pairings improve decisions
on six tasks. Each history contains 128 records.
DPDM\_Qwen3.5-27B uses the original input--target pairs.
Permuted historical labels retain the same inputs and class counts but
shuffle their target assignments. The most-frequent-label rule always
returns the modal historical target without conditioning on the query.}
\label{tab:history_pairing}
\small
\setlength{\tabcolsep}{4pt}
\renewcommand{\arraystretch}{1.05}
\begin{tabularx}{\columnwidth}{@{}Xr@{}}
\toprule
Condition & Accuracy (\%) \\
\midrule
DPDM\_Qwen3.5-27B & 80.71 $\pm$ 0.41 \\
Permuted historical labels & 50.13 $\pm$ 0.80 \\
Most-frequent-label rule & 50.52 $\pm$ 1.50 \\
\bottomrule
\end{tabularx}
\end{table}

DPDM reaches 80.71\% accuracy, compared with 50.13\%
after permuting the historical labels (Table~\ref{tab:history_pairing}).
Always choosing the most frequent historical label scores 50.52\%.
The original labels give higher accuracy on all six tasks, even though
their class frequencies are unchanged by permutation.

With the original input--target pairings preserved, we compare adding
new observations with repeating examples already in the history.
The three conditions contain 16 distinct records, those
same records repeated eight times, or 128 distinct records. This
comparison uses the same six
tasks and 737 queries as Table~\ref{tab:history_pairing}. All conditions
share a feature map fitted on the 128 unlabeled historical inputs;
only the labeled records supplied to the predictor change.

\begin{table}[!htbp]
\centering
\caption{Distinct observations provide more useful history than
repeated examples on the same six tasks.}
\label{tab:history_records}
\small
\setlength{\tabcolsep}{3pt}
\renewcommand{\arraystretch}{1.05}
\begin{tabularx}{\columnwidth}{@{}Xrr@{}}
\toprule
Historical records & Rows & Accuracy (\%) \\
\midrule
16 distinct & 16 & 60.31 $\pm$ 3.49 \\
16 repeated 8 times & 128 & 64.79 $\pm$ 2.02 \\
128 distinct (DPDM) & 128 & 80.71 $\pm$ 0.41 \\
\bottomrule
\end{tabularx}
\end{table}

Repetition raises accuracy from 60.31\% to 64.79\%, while 128 distinct
observations reach 80.71\% (Table~\ref{tab:history_records}). At the same
128-row size, distinct observations yield an absolute accuracy gain of
15.92\% over repetition, with gains on all six tasks.

To assess the contribution of semantic features, we remove them while
leaving the native and lexical features unchanged. This ablation
expands the six-task set with Emotion and WANLI
and uses all 18,759 queries from the eight tasks. At each history size,
DPDM and its ablations use identical historical inputs
and queries. A second ablation retains the full representation but
permutes the historical labels. Each task uses its largest available
history within each of the four ceilings in Table~\ref{tab:semantic_ablation}.

\begin{table}[!htbp]
\centering
\caption{Accuracy (\%) on eight tasks with semantic-feature and
historical-label ablations.}
\label{tab:semantic_ablation}
\small
\setlength{\tabcolsep}{3pt}
\renewcommand{\arraystretch}{1.12}
\begin{tabular}{@{}rrrr@{}}
\toprule
\shortstack{History\\ceiling} & DPDM &
\shortstack{No semantic\\features} & \shortstack{Permuted\\labels} \\
\midrule
16 & 58.20 $\pm$ 2.55 & 43.04 $\pm$ 2.88 & 38.93 $\pm$ 3.09 \\
32 & 67.80 $\pm$ 1.09 & 51.63 $\pm$ 2.08 & 42.95 $\pm$ 2.00 \\
128 & 75.90 $\pm$ 0.44 & 57.21 $\pm$ 1.63 & 46.45 $\pm$ 1.06 \\
512 & 80.42 $\pm$ 0.33 & 63.95 $\pm$ 0.31 & 48.78 $\pm$ 1.99 \\
\bottomrule
\end{tabular}
\end{table}

At a ceiling of 512 records,
removing semantic features lowers accuracy from 80.42\% to 63.95\%;
permuting labels lowers it to 48.78\%. DPDM has higher
mean accuracy at every ceiling and outperforms both ablations on all
eight tasks at the highest ceiling.
These controls show that semantic features and historical input--target
relationships both contribute to decision accuracy, while distinct
observations provide information beyond repeated examples.
Together with the encoder and matched-history comparisons in
Section~\ref{sec:exp_generality}, these findings support two complementary
routes to improvement: better semantic representations and more effective
use of historical feedback.
Appendix~\ref{app:qwen_controls} gives the sampling procedures
and history budgets for these comparisons.

%% file: sections/cost_current.tex
\begin{figure}[!htbp]
\centering
\includegraphics[width=0.78\columnwidth]{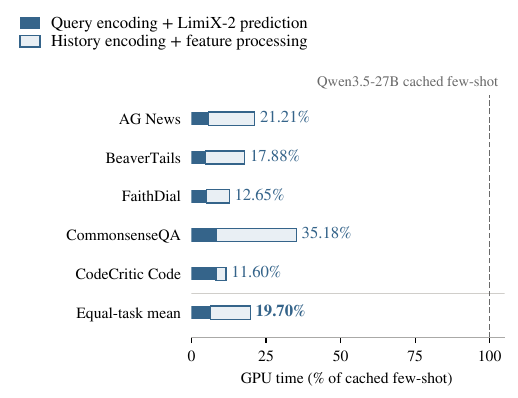}
\caption{Measured GPU time relative to cached few-shot prompting.
Both totals include historical preparation and processing of the same
256 queries per task. The last row gives the equal-task arithmetic mean.}
\label{fig:efficiency}
\end{figure}

\paragraph{Computational cost.}
DPDM reuses encoded historical inputs across subsequent
queries, reducing repeated semantic processing of the history.
Figure~\ref{fig:efficiency} compares measured GPU time for Qwen3.5-27B
cached few-shot prompting and DPDM on five decision tasks,
using identical histories and 256 queries per task. Both totals include
historical preparation and query processing. DPDM's full-pipeline
GPU time averages 19.7\% of the cached few-shot baseline, a reduction
of 80.3\%, with the five taskwise cost ratios weighted equally.
Appendix~\ref{app:cost} gives the timing protocol.

%% file: sections/discussion_conclusion.tex
\section{Discussion and Conclusion}
\subsection{Discussion}
\label{sec:discussion}

\paragraph{Learning from experience as a foundation-model capability.}
The central implication is architectural: an agent's capacity to learn
from deployment experience can be developed through a pretrained model
dedicated to conditional prediction. Language pretraining supplies knowledge
for interpreting requests, contexts, and candidate choices. Conditional
pretraining supplies a mechanism for learning relationships from the
records available in a particular deployment. Connecting these capabilities
allows semantic understanding and empirical learning to contribute to the
same decision. The value of the separation is that it gives learning from
experience an explicit role in agent design, with its own training
objectives and evaluation criteria.

\paragraph{Relation to prior work.}
This proposal brings together two research directions. ExpeL extracts
insights from experience and retrieves examples to guide an LLM
agent~\cite{zhao2023expel}. DPT learns to predict optimal actions from
in-context interaction data~\cite{lee2023dpt}, while Prior-Data Fitted
Networks pretrain conditional predictors across sampled prediction
problems~\cite{muller2022bayesian}. ICR-RL and BC-ICL apply general
predictive foundation models to control and contextual-bandit
decisions~\cite{schiff2026icrrl,gupta2026bcicl}. DPDM connects
conditional predictors to semantically expressed decision tasks through
language-model representations. Each task retains its own meaning and
target, while the learning mechanism operates on represented records and
their labels. The use of LimiX-2 across semantic encoders illustrates how
this organization can accommodate different ways of representing a
problem. It also creates a common interface for evaluating future conditional
predictors with the same semantic encoders.

\paragraph{What makes experience valuable.}
The information--inference distinction clarifies how this architecture
can improve. New observations can reveal relationships that the existing
history leaves uncertain; improving the learner can make more effective
use of relationships already supported by that history. These are
different research goals. A record's value therefore depends on what it
adds about the current decision and on the mechanism used to interpret it.
The history-growth and imbalanced-history analyses give concrete examples
of this principle. They motivate studying how agents collect and retain
feedback alongside how conditional predictors are pretrained to learn from it.
Under this view, deployment history helps determine the decision function
instantiated for the current setting.

\subsection{Conclusion}
\label{sec:conclusion}

We propose an architecture for agent decision making in which semantic
understanding and learning from experience are distinct, connected
foundation-model capabilities. By expressing task-defined decisions as
conditional prediction problems, language-model representations allow a
shared conditional predictor to use deployment histories across
heterogeneous tasks. The LimiX-2 instantiation demonstrates gains across
semantic encoders, growing histories, and imbalanced feedback.
This architecture provides a common interface through which advances
in learning from experience can improve decisions across agent tasks.

%% file: sections/method_appendix.tex
\section{Experimental Instantiation}
\label{app:implementation}

This section describes the evaluated DPDM implementation, which pairs a frozen
semantic encoder with LimiX-2. The lexical features, native fields, and
dimensionality-reduction methods described below specify this implementation.

\subsection{Task Records and Historical Feedback}

For each task, a record contains the visible input fields and candidate
descriptions. The task specification defines the question to be answered and
the mapping between answers and labels. For example, BFCL Callability supplies
a user request and function schemas; the target indicates whether any supplied
function is applicable. The same specification is used for historical records
and queries. Each record becomes one feature row. LimiX-2 receives the historical
rows, their observed labels, and the query row whose label is to be predicted.
Labels and source-group identifiers are excluded from the record features.

Where source groups are defined, history and queries are separated by group.
Numerical history budgets use nested prefixes of a randomly permuted historical
pool, sampled without replacement or class balancing. Full-pool conditions
retain the original record order. Appendix~\ref{app:exp_protocol} gives the
evaluation budgets and sampling settings.

\subsection{Record Representations}

Each record is encoded independently using its task question and candidate
descriptions. Qwen and Llama use their native chat templates with text-only
inputs; Qwen thinking is disabled. The feature row combines three components:

\begin{itemize}
\item \textbf{Semantic features.} We read the final post-normalization hidden
state at the last nonpadding prompt token, before answer generation. States
from multiple target prompts are concatenated, L2-normalized, and compressed
by PCA to at most 32 dimensions.
\item \textbf{Lexical features.} TF-IDF features~\cite{salton1988term} represent
the visible text fields, with history-fitted dimensionality reduction using
truncated SVD when applicable~\cite{deerwester1990indexing}.
\item \textbf{Native fields.} Available numeric and categorical fields accompany
the text features. Categorical mappings and numeric scaling are fitted on the
historical inputs; unseen query categories are treated as missing values.
\end{itemize}

Feature transformations are fitted only on historical inputs and applied
unchanged to queries. As history grows, these transformations are refitted while
both neural models remain frozen. Encoded states can be reused when the encoder,
prompt renderer, and task specification are unchanged. The controls in
Appendix~\ref{app:qwen_controls} specify when the fitted transformation is held
fixed across conditions.

\subsection{From Predictions to Decisions}

All tasks use the same frozen LimiX-2 model with 32-pipeline aggregation.
Prediction uses all supplied historical rows, with retrieval disabled. LimiX-2 estimates label
probabilities from these rows and the current query. For more than ten observed
classes, a deterministic label hierarchy combines parent and leaf probabilities
to recover probabilities for the original task labels.

Classes absent from history initially receive zero probability. A history
containing one observed class produces a probability of one for that class
without invoking LimiX-2. Candidates unavailable for the current query are then
removed and the remaining probabilities normalized. If no probability mass
remains, the valid candidates receive equal probabilities. The candidate with
the highest probability is selected, with ties resolved by candidate order.

The controlled outcome example supplies numeric context--action inputs directly
to LimiX-2 and selects the action with the highest predicted success probability.

%% file: sections/experiments_appendix.tex
\section{Benchmark Inventory}
\label{app:inventory}

The evaluation covers 50 benchmarks across five decision categories.
\input{sections/benchmark_guide}

\par\medskip\noindent
\begin{minipage}{\columnwidth}
\centering
\captionsetup{hypcap=false}
\captionof{table}{Benchmark inventory, part I. Histories count labeled records;
query counts sum task instances, which can share source examples.}
\label{tab:inventory}
\scriptsize
\setlength{\tabcolsep}{2pt}
\resizebox{\linewidth}{!}{\input{tables/inventory_one}}
\end{minipage}
\par\medskip\noindent
\begin{minipage}{\columnwidth}
\centering
\captionsetup{hypcap=false}
\captionof{table}{Benchmark inventory, part II. Definitions follow
Table~\ref{tab:inventory}.}
\label{tab:inventory_two}
\scriptsize
\setlength{\tabcolsep}{2pt}
\resizebox{\linewidth}{!}{\input{tables/inventory_two}}
\end{minipage}
\par\medskip

\FloatBarrier
\section{Evaluation Details}
\label{app:exp_protocol}

\subsection{Prompting, History Budgets, and Aggregation}

The main comparison uses 82 task configurations. DPDM's task-specific
history budget is the largest completed size shared by the four main
encoders, the four supplementary StartLux encoders, and three history
samples. History-sampling seeds are 11, 23, and 47 for LimiX-2 and 11 for
prompting. Each condition evaluates the complete task query set.
We average task configurations within each dataset, then weight datasets
equally. DPDM's mean and sample standard deviation are computed from
the three resulting dataset-macro scores. Table~\ref{tab:budget_ranges}
summarizes the history ranges.

Prompt histories are evaluated from larger to smaller sizes subject to
each model's context window, GPU-memory availability, and per-condition
runtime budget. The reported endpoint is the largest
completed condition, chosen independently of its accuracy.

\begin{table}[!htbp]
\centering
\caption{History ranges for the main and supplementary comparisons,
measured in labeled records.}
\label{tab:budget_ranges}
\scriptsize
\input{tables/budget_ranges}
\end{table}

Qwen and Llama prompting uses native chat templates and candidate-letter
next-token probabilities; Qwen thinking is disabled. StartLux uses its
native decision readout. Large candidate sets are grouped; group and within-group probabilities
are multiplied to score each candidate. Jev API v1.13.0 uses the official service with
compact candidate representations and its largest completed history
for each task, subject to the service's call and context limits.
Its zero-shot and few-shot conditions share query IDs, labels, task
definitions, and service version.

The matched-history comparison in Section~\ref{sec:exp_generality}
uses 82 configurations from 50 datasets. Within each encoder--task pair,
both methods receive the same ordered historical records, labels, and
queries at the encoder's largest completed common prompt history,
using seed 11.

\subsection{History-Growth Evaluation}
\label{app:task_curves}

The curves in Figure~\ref{fig:qwen_history} use all task configurations
with complete coverage across eight encoders, five budgets
(32, 128, 512, 2,048, and 8,192), and three sampling seeds, with unchanged
query IDs, within the main-evaluation tasks.
This yields 21 datasets and 29 task configurations.
The datasets are ADE Corpus, AG News, ANLI, BeaverTails, BoolQ, Circa,
CoLA, CommonsenseQA, CosmosQA, Emo, FaithDial, Google Well-formed Query,
HH-RLHF, Hate Speech / Offensive, HateSpeech18, HateXplain, Health Fact,
HellaSwag, PAWS, SNLI, and TweetEval Sentiment.

Figure~\ref{fig:task_scaling} uses DPDM with Qwen3.5-27B as the semantic
encoder for four agent-related tasks. Table~\ref{tab:task_history_budgets} gives the endpoint
history sizes and prompt-reference budgets. All four DPDM endpoints
use the full historical pool in its original order, shared by all three
history samples; preceding numerical budgets use sampled prefixes.
The figure displays histories of at least 16 records. The displayed Qwen
few-shot curves use $M=16,32,64$ for BFCL and CodeCritic QA, and
$M=16,32,64,128$ for ToolSandbox and CodeCritic Code, with seed 11
and the unchanged query sets. The table lists the largest completed
Qwen few-shot history and the Jev reference history for each task.
ToolPRMBench BFCL evaluates tool-use steps, as distinct from the
tool-applicability judgments in BFCL Callability.

\begin{table}[!htbp]
\centering
\caption{Historical record counts for Figure~\ref{fig:task_scaling}.}
\label{tab:task_history_budgets}
\footnotesize
\setlength{\tabcolsep}{3pt}
\begin{tabular}{@{}lrrr@{}}
\toprule
Task & DPDM & Qwen & Jev \\
 & endpoint & few-shot & few-shot \\
\midrule
ToolPRMBench BFCL & 233 & 64 & 16 \\
ToolPRMBench ToolSandbox & 228 & 128 & 16 \\
CodeCritic Code & 2,238 & 128 & 32 \\
CodeCritic QA & 773 & 64 & 16 \\
\bottomrule
\end{tabular}
\end{table}

\input{sections/supplementary_models}
\input{sections/history_bias_details}
\FloatBarrier
\input{sections/qwen_control_details}
\FloatBarrier
\input{sections/cost_details}

%% file: sections/benchmark_guide.tex
Each benchmark defines a concrete decision and its reference target.
DPDM uses historical feedback to predict that target for the
current input. The 82 task configurations span five categories:

\begin{itemize}
\item \textbf{Agent tasks.}
Inputs include requests, available tools, reasoning steps, or candidate
responses. Targets specify an intent, tool applicability, response type,
step assessment, or preferred response. This category includes Banking77,
TREC, Google Well-formed Query, BFCL Callability, ToolACE, ToolPRMBench,
AgentProcess, CodeCritic, and HH-RLHF.
\item \textbf{Knowledge QA.}
Tasks include answer selection and reasoning-step judgments from
questions, passages, and candidate answers or steps.
Examples include ARC, CommonsenseQA, HellaSwag, HaluEval,
PRM800K, and BIG-Bench Mistake.
\item \textbf{Evidence consistency.}
Claims, responses, or sentence pairs are judged against their supplied
context. Examples include FaithBench, RAGTruth, and natural-language
inference tasks.
\item \textbf{Safety and reliability.}
Messages, responses, or reports are assigned task-defined risk or reliability labels.
Examples include BeaverTails, ToxicChat, hate-speech datasets, ADE Corpus,
and FaithDial's dialogue-response hallucination judgments.
\item \textbf{Language understanding.}
Text is assigned a topic, sentiment, emotion, or linguistic label.
Examples include AG News, TweetEval Sentiment, Emo, and grammatical-acceptability tasks.
\end{itemize}

BFCL Callability asks whether any supplied function applies to a user
request~\cite{patil2025bfcl}. The response-mode task derived from ToolACE
predicts whether the next response is text, one tool call, or multiple
calls; its targets come from
the response structure in the source records~\cite{liu2024toolace}.
ToolPRMBench's BFCL, ToolSandbox, and ToolTalk suites evaluate judgments
about tool-use steps~\cite{li2026toolprmbench}. Thus BFCL Callability and
ToolPRMBench-BFCL have different inputs and targets.
BIG-Bench Mistake supplies a question, preceding reasoning, and the
current step, whose correctness is the target~\cite{tyen2024mistake}.
CodeCritic includes Code and QA assessment tasks~\cite{zhang2025codecritic}.
PRM800K evaluates reasoning steps~\cite{lightman2023verify}.
The scores in this paper measure these individual labeled judgments.

Each task supplies a query set, a separate historical pool, and a target
definition. Appendix~\ref{app:implementation} describes record representations,
and Appendix~\ref{app:exp_protocol} gives the evaluation and
history-sampling protocols.
Tables~\ref{tab:inventory} and \ref{tab:inventory_two} give the dataset
inventory and history ranges.

%% file: tables/inventory_one.tex
\begin{tabular}{@{}lrrrr@{}}
\toprule
Dataset & Tasks & Queries & LM history & DPDM history \\
\midrule
ADE Corpus & 1 & 4,704 & 1,024--4,096 & 8,192--8,192 \\
AG News & 2 & 15,200 & 512--2,048 & 2,048--8,192 \\
ANLI & 4 & 4,200 & 1,024--2,048 & 8,192--8,192 \\
ARC & 2 & 1,471 & 1,118--1,118 & 1,024--1,024 \\
AgentProcess & 1 & 2,351 & 24--96 & 267--267 \\
BFCL Callability & 1 & 696 & 128--256 & 2,175--2,175 \\
BIG-Bench Mistake & 5 & 7,032 & 128--1,024 & 512--4,096 \\
BLiMP & 4 & 800 & 800--800 & 512--512 \\
Banking77 & 1 & 3,080 & 77--77 & 4,928--4,928 \\
BeaverTails & 2 & 36,417 & 32--2,048 & 8,192--8,192 \\
BoolQ & 2 & 6,540 & 256--1,024 & 8,192--8,192 \\
CB & 1 & 56 & 32--32 & 128--128 \\
CODAH & 1 & 556 & 1,665--1,665 & 1,024--1,024 \\
Circa & 1 & 6,700 & 512--4,096 & 8,192--8,192 \\
Climate-FEVER & 1 & 307 & 1,228--1,228 & 1,024--1,024 \\
CoLA & 2 & 2,086 & 2,048--8,192 & 8,192--8,192 \\
CodeCritic & 2 & 1,287 & 64--256 & 773--2,238 \\
CommonsenseQA & 1 & 1,221 & 1,024--4,096 & 8,192--8,192 \\
CosmosQA & 1 & 2,985 & 512--1,024 & 8,192--8,192 \\
Emo & 1 & 5,509 & 1,024--4,096 & 8,192--8,192 \\
Ethos & 7 & 609 & 346--346 & 256--256 \\
FaithBench & 1 & 310 & 128--440 & 256--256 \\
FaithDial & 1 & 3,539 & 256--1,024 & 8,192--8,192 \\
Financial PhraseBank & 1 & 970 & 2,048--3,868 & 2,048--2,048 \\
Google Well-formed Query & 1 & 3,297 & 2,048--8,192 & 8,192--8,192 \\
\bottomrule
\end{tabular}

%% file: tables/inventory_two.tex
\begin{tabular}{@{}lrrrr@{}}
\toprule
Dataset & Tasks & Queries & LM history & DPDM history \\
\midrule
HH-RLHF & 2 & 8,499 & 128--1,024 & 8,192--8,192 \\
HaluEval & 1 & 4,000 & 256--1,024 & 4,096--4,096 \\
Hate Speech / Offensive & 1 & 4,957 & 1,024--4,096 & 8,192--8,192 \\
HateSpeech18 & 1 & 2,141 & 2,048--4,096 & 8,192--8,192 \\
HateXplain & 1 & 1,922 & 2,048--4,096 & 8,192--8,192 \\
Health Fact & 1 & 1,214 & 2,048--4,096 & 8,192--8,192 \\
HellaSwag & 2 & 20,084 & 128--1,024 & 8,192--8,192 \\
MRPC & 2 & 2,133 & 1,024--2,048 & 2,048--2,048 \\
PAWS & 1 & 8,000 & 512--2,048 & 8,192--8,192 \\
PRM800K & 1 & 5,965 & 96--384 & 1,536--1,536 \\
ProofWriter & 2 & 11,732 & 48--384 & 1,536--1,536 \\
PubMedQA & 1 & 500 & 256--500 & 256--256 \\
RAGTruth & 3 & 2,693 & 64--256 & 1,496--4,974 \\
SMS Spam & 1 & 2,197 & 2,048--3,104 & 2,048--2,048 \\
SNLI & 1 & 9,824 & 512--4,096 & 8,192--8,192 \\
SemEval-2010 Task 8 & 1 & 2,717 & 256--2,048 & 4,096--4,096 \\
TREC & 2 & 1,000 & 256--4,839 & 4,096--4,096 \\
ToolACE & 1 & 1,380 & 96--192 & 1,536--1,536 \\
ToolPRMBench & 3 & 233 & 32--162 & 58--233 \\
ToxicChat & 2 & 10,166 & 512--4,096 & 4,096--4,096 \\
TweetEval Sentiment & 1 & 12,284 & 512--4,096 & 8,192--8,192 \\
WANLI & 1 & 5,000 & 768--1,536 & 1,536--1,536 \\
WNLI & 1 & 71 & 635--635 & 512--512 \\
WSC & 1 & 104 & 552--552 & 512--512 \\
WiC & 1 & 638 & 1,024--2,048 & 4,096--4,096 \\
\bottomrule
\end{tabular}

%% file: tables/budget_ranges.tex
\begin{tabular}{@{}lrr@{}}
\toprule
Encoder/service & Prompt records & DPDM records \\
\midrule
StartLux-0.8B & 32--4,839 & 58--8,192 \\
StartLux-4B & 24--4,839 & 58--8,192 \\
StartLux-9B & 32--4,839 & 58--8,192 \\
StartLux-27B & 24--2,048 & 58--8,192 \\
Qwen3.5-4B & 32--8,192 & 58--8,192 \\
Qwen3.5-9B & 32--8,192 & 58--8,192 \\
Qwen3.5-27B & 24--4,839 & 58--8,192 \\
Llama-3.1-8B & 24--4,096 & 58--8,192 \\
Jev API & 6--77 & --- \\
\bottomrule
\end{tabular}

%% file: sections/supplementary_models.tex
\section{Learning from History after Decision-Specific Post-Training}
\label{app:extended_results}

StartLux tests whether conditional prediction can further improve a
language model already trained for decision tasks. Its technical report
describes full-parameter supervised post-training on a large collection
of decision examples~\cite{startlux2026}. We evaluate the original
StartLux v1 checkpoints at 0.8B, 4B, 9B, and 27B~\cite{startluxv1},
both through their native decision readout and as frozen semantic
encoders for the same LimiX-2 checkpoint.

\begin{table}[!htbp]
\centering
\caption{Decision accuracy (\%) with StartLux.
The final column pairs each frozen StartLux encoder with LimiX-2.}
\label{tab:specialized_comparison}
\small
\setlength{\tabcolsep}{4pt}
\input{tables/startlux_supplement}
\end{table}

Table~\ref{tab:specialized_comparison} uses the 50-dataset evaluation
and aggregation settings in Appendix~\ref{app:exp_protocol}.
At 27B, few-shot accuracy is 83.85\%; using the same frozen model as
DPDM's semantic encoder raises it to 86.57\%. DPDM also improves on
both prompting settings at the other three model sizes.
On the matched histories used in Section~\ref{sec:exp_generality},
StartLux-27B improves from 83.85\% to 84.20\%.

These results show that the decision capability acquired through
large-scale post-training leaves room for further gains from historical
feedback. DPDM supplies that gain through conditional prediction,
with both models' parameters fixed.

%% file: tables/startlux_supplement.tex
\begin{tabular}{@{}lrrr@{}}
\toprule
Encoder & Zero-shot & Few-shot & DPDM \\
\midrule
StartLux-0.8B & 65.19 & 54.91 & \textbf{78.80} $\pm$ 0.04 \\
StartLux-4B & 72.32 & 72.97 & \textbf{84.06} $\pm$ 0.02 \\
StartLux-9B & 74.92 & 75.20 & \textbf{85.25} $\pm$ 0.14 \\
StartLux-27B & 78.66 & 83.85 & \textbf{86.57} $\pm$ 0.11 \\
\bottomrule
\end{tabular}

%% file: sections/history_bias_details.tex
\section{Imbalanced-History Details}
\label{app:history_bias}

This section gives the encoder--task results underlying
Figure~\ref{fig:history_bias}. The comparison uses all five BIG-Bench
Mistake tasks, the three Qwen3.5 encoders, and Llama-3.1-8B.
The 7,032 query steps comprise 6,347 correct and 685 incorrect steps.
The task-level correct/incorrect counts are 4,805/270 for Dyck languages,
270/120 for logical deduction, 363/88 for arithmetic, 358/104 for object
tracking, and 551/103 for word sorting. Historical and query reasoning
traces are separated by their source-group identifiers.

Each encoder--task pair uses seed 11 and its largest completed history
with identical ordered history and query IDs for few-shot prompting
and DPDM using the same encoder. The common histories contain 128--1,024 records, with
66.41--95.31\% correct steps.

Table~\ref{tab:history_bias_task_scores} reports ordinary accuracy,
balanced accuracy, and the accuracy within each reference class.
The correct-step score is the fraction of correct steps accepted;
the incorrect-step score is the fraction of incorrect steps rejected.
Balanced accuracy is their mean. Figure~\ref{fig:history_bias}
averages the four encoders within each task and gives the five tasks
equal weight in its overall mean.

\begin{table*}[!t]
\centering
\caption{Encoder--task details for the imbalanced-history analysis.
$M$ counts historical records; Hist.\ correct is their correct-step
percentage. Acc.\ and BA denote ordinary and balanced accuracy.
$A_{\mathrm{correct}}$ and $A_{\mathrm{incorrect}}$ are class-specific
accuracies. All scores are percentages.}
\label{tab:history_bias_task_scores}
\footnotesize
\setlength{\tabcolsep}{3.5pt}
\input{tables/history_bias_main_details}
\end{table*}

%% file: tables/history_bias_main_details.tex
\begin{tabular}{@{}lrrrrrrrrrr@{}}
\toprule
& & & \multicolumn{4}{c}{Few-shot} & \multicolumn{4}{c}{DPDM} \\
\cmidrule(lr){4-7}\cmidrule(lr){8-11}
Encoder & $M$ & Hist.\ correct & Acc. & BA & $A_{\mathrm{correct}}$ & $A_{\mathrm{incorrect}}$ & Acc. & BA & $A_{\mathrm{correct}}$ & $A_{\mathrm{incorrect}}$ \\
\midrule
\multicolumn{11}{l}{\textit{Dyck languages}} \\
Qwen3.5-4B & 1024 & 94.92 & 90.74 & 54.04 & 95.11 & 12.96 & 96.00 & 62.76 & 99.96 & 25.56 \\
Qwen3.5-9B & 256 & 95.31 & 94.88 & 54.82 & 99.65 & 10.00 & 92.45 & 52.84 & 97.17 & 8.52 \\
Qwen3.5-27B & 1024 & 94.92 & 95.05 & 53.52 & 100.00 & 7.04 & 95.68 & 60.32 & 99.90 & 20.74 \\
Llama-3.1-8B & 256 & 95.31 & 94.09 & 49.86 & 99.35 & 0.37 & 93.54 & 52.19 & 98.46 & 5.93 \\
\midrule
\multicolumn{11}{l}{\textit{Logical deduction}} \\
Qwen3.5-4B & 571 & 69.53 & 69.23 & 51.62 & 97.41 & 5.83 & 80.77 & 85.88 & 72.59 & 99.17 \\
Qwen3.5-9B & 571 & 69.53 & 67.95 & 76.16 & 54.81 & 97.50 & 86.15 & 88.84 & 81.85 & 95.83 \\
Qwen3.5-27B & 256 & 66.41 & 93.33 & 89.86 & 98.89 & 80.83 & 94.62 & 93.33 & 96.67 & 90.00 \\
Llama-3.1-8B & 256 & 66.41 & 69.49 & 50.42 & 100.00 & 0.83 & 83.08 & 82.92 & 83.33 & 82.50 \\
\midrule
\multicolumn{11}{l}{\textit{Multistep arithmetic}} \\
Qwen3.5-4B & 642 & 76.64 & 71.18 & 60.14 & 78.24 & 42.05 & 51.88 & 69.68 & 40.50 & 98.86 \\
Qwen3.5-9B & 642 & 76.64 & 76.72 & 64.45 & 84.57 & 44.32 & 49.22 & 68.03 & 37.19 & 98.86 \\
Qwen3.5-27B & 642 & 76.64 & 76.72 & 67.89 & 82.37 & 53.41 & 64.97 & 76.95 & 57.30 & 96.59 \\
Llama-3.1-8B & 642 & 76.64 & 80.49 & 50.00 & 100.00 & 0.00 & 47.89 & 67.63 & 35.26 & 100.00 \\
\midrule
\multicolumn{11}{l}{\textit{Object tracking}} \\
Qwen3.5-4B & 689 & 77.36 & 51.08 & 51.72 & 50.56 & 52.88 & 91.56 & 89.44 & 93.30 & 85.58 \\
Qwen3.5-9B & 689 & 77.36 & 75.54 & 73.64 & 77.09 & 70.19 & 91.56 & 89.78 & 93.02 & 86.54 \\
Qwen3.5-27B & 689 & 77.36 & 84.85 & 69.07 & 97.77 & 40.38 & 94.37 & 92.62 & 95.81 & 89.42 \\
Llama-3.1-8B & 512 & 78.32 & 77.49 & 50.00 & 100.00 & 0.00 & 90.04 & 84.02 & 94.97 & 73.08 \\
\midrule
\multicolumn{11}{l}{\textit{Word sorting}} \\
Qwen3.5-4B & 512 & 81.84 & 85.32 & 57.74 & 98.00 & 17.48 & 62.39 & 72.94 & 57.53 & 88.35 \\
Qwen3.5-9B & 512 & 81.84 & 15.75 & 50.00 & 0.00 & 100.00 & 66.06 & 74.72 & 62.07 & 87.38 \\
Qwen3.5-27B & 128 & 85.94 & 90.52 & 72.27 & 98.91 & 45.63 & 89.45 & 72.03 & 97.46 & 46.60 \\
Llama-3.1-8B & 256 & 83.20 & 84.25 & 50.00 & 100.00 & 0.00 & 84.25 & 50.39 & 99.82 & 0.97 \\
\bottomrule
\end{tabular}

%% file: sections/qwen_control_details.tex
\section{History and Representation Controls}
\label{app:qwen_controls}

The Section~\ref{sec:exp_controls} controls use frozen Qwen3.5-27B
representations and LimiX-2 with 32-pipeline aggregation.
History-sampling seeds are 11, 23, and 47.

\paragraph{Historical-record controls.}
BFCL Callability, ToolPRMBench ToolSandbox, CodeCritic Code, PRM800K,
FaithDial, and TREC coarse each supply 128 historical records per seed.
Query selection is independent of labels: a fixed random sample contains
128 queries per task, except ToolSandbox, which contributes all 97.
This gives 737 query instances per seed, shared by every condition.
Historical and query IDs are disjoint.

For each task and seed, the feature map is fitted once on the 128
unlabeled historical inputs and then reused. Label permutation selects
0, 13, 32, 64, or 128 historical positions and permutes their labels
within the selected set. The position sets are nested. Class counts
remain exactly unchanged; some selected positions can keep their
original label by chance. The recorded outputs distinguish the selected
fraction from the fraction whose label actually changes.

The distinct-record comparison supplies the first 16 labeled records,
those records repeated eight times, or all 128 distinct labeled records.
It uses the same fitted representation in each case. Thus, the 16-record
condition has 16 labeled examples and a transformation fitted from the
same 128 unlabeled reference inputs as the other conditions. Repetition
changes row multiplicity without adding a new labeled observation.
The historical-frequency reference predicts the most common label in
the 128-record history for every query, with ties resolved by label ID.

The representation controls evaluate the complete query sets of the eight
tasks, totaling 18,759 queries. The tasks are the six above plus Emotion
and WANLI, with 696, 97, 962, 5,965, 3,539, 500, 2,000, and 5,000 queries,
respectively. We reuse the original history and query IDs at each task's
largest numerical history within ceilings of 16, 32, 128, and 512. The
corresponding actual history sizes range from 12--16, 24--32, 96--128,
and 162--512 records.

Each task--history--seed group compares the full representation,
the same representation with semantic feature columns removed, and
the full representation with permuted historical labels. Native and
lexical features are unchanged by semantic-feature removal. All
variants use the same queries and history-fitted transformations.

\paragraph{Aggregation and reported contrasts.}
We average task accuracies equally within each seed, then report the
mean and sample standard deviation across the three seeds. Main-text
contrasts compare original with permuted labels, distinct with repeated
records at equal row count, and full with semantic-feature-removed
representations at the same history size. The six-task and eight-task
means are reported separately in Tables~\ref{tab:history_pairing},
\ref{tab:history_records}, and~\ref{tab:semantic_ablation}.

%% file: sections/cost_details.tex
\section{Computational Cost}
\label{app:cost}

The five tasks are AG News, BeaverTails, FaithDial, CommonsenseQA,
and CodeCritic Code. Both pipelines use Qwen3.5-27B on dedicated Pro6000
GPUs. Within each task, they receive the same historical records at
the largest completed Qwen3.5-27B few-shot history size from the main
evaluation and the first 256 queries in the original test order.
Table~\ref{tab:cost_cases} gives the history sizes and measured totals.

The DPDM total includes historical encoding, feature processing,
query encoding, and full 32-pipeline LimiX-2 prediction. Feature processing
covers fitting the transformation on historical inputs and applying it
to query inputs. The cached few-shot total includes one prefix-cache
construction per task and inference on all 256 queries. Its query times
sum four 64-query blocks per repetition; all queries reuse the prefix
cache. The baseline processes queries individually.

We report the sum of measured component times on an occupied GPU,
including historical preparation and query processing. Historical encoding averages
three runs for CodeCritic Code and uses one run for each other task;
all other components average three runs. Model loading, separate warm-up
runs, and validation are excluded; overhead within retained measurements
is included.

For each task, the reported ratio is
\[
100\,\frac{T_{\mathrm{framework}}}{T_{\mathrm{cached}}}.
\]
Both terms denote full-pipeline GPU time, including each method's
task-specific historical preparation. We report the arithmetic mean
of the five unrounded taskwise ratios, with equal weight for each task.

\begin{table}[!htb]
\centering
\caption{Full-pipeline GPU-hours on five tasks, with 256 queries per task.
Both methods include historical preparation. $M$ counts historical records.}
\label{tab:cost_cases}
\small
\setlength{\tabcolsep}{3pt}
\begin{tabular}{lrrr}
\toprule
Task & $M$ & DPDM & Cached few-shot \\
 & & GPU-h & GPU-h \\
\midrule
AG News & 1,024 & 0.036234 & 0.170823 \\
BeaverTails & 1,024 & 0.036845 & 0.206056 \\
FaithDial & 512 & 0.023341 & 0.184489 \\
CommonsenseQA & 2,048 & 0.072727 & 0.206743 \\
CodeCritic Code & 128 & 0.021397 & 0.184467 \\
\bottomrule
\end{tabular}
\end{table}

%% file: sections/decision_quality.tex
\section{Information and Inference in Decision Quality}
\label{sec:decision_value}

Additional observations can reveal a task relationship; a better learner
can estimate it more accurately from those observations. To separate
these effects, fix a task
distribution and loss $L(b,Y)$ for answer $b$ and target $Y$.
Risk is the expected loss, such as the error rate under zero--one loss.
Let $I$ contain the represented support and query, historical targets,
and task specification. The best risk achievable using this information is
\begin{equation}
 R_I^\star=\inf_{\pi\in\Pi_I}\mathbb E[L(\pi(I),Y)],
 \label{eq:reference_risk}
\end{equation}
where $\Pi_I$ contains legal decision rules based only on $I$.
Let $R_m$ be the actual learner's risk. An oracle receives a richer
information set $I_{\mathrm{or}}$ containing $I$, giving reference risk
$R_{I_{\mathrm{or}}}^\star$. The learner's excess risk splits into
\begin{equation}
 \begin{aligned}
 R_m-R_{I_{\mathrm{or}}}^\star
 ={}&\underbrace{R_I^\star-R_{I_{\mathrm{or}}}^\star}_{\text{information gap}}\\
 &+\underbrace{R_m-R_I^\star}_{\text{inference gap}}.
 \end{aligned}
 \label{eq:gap_decomposition}
\end{equation}
The information gap is the loss that even the best rule using $I$ cannot
avoid without the oracle's additional information. The inference gap is
the learner's loss beyond that best available-information rule.
Both are nonnegative under the fixed distribution, loss, and nested
information sets.

Appendix~\ref{app:decision_theory} illustrates this decomposition in a
controlled environment with known success probabilities. An oracle knows
the deployment mechanism, and a Bayes rule infers it from history;
these references make the two gaps computable.

%% file: sections/decision_theory.tex
\FloatBarrier
\section{A Controlled Example}
\label{app:decision_theory}
\label{app:controlled_protocol}

The decomposition in Appendix~\ref{sec:decision_value} distinguishes
uncertainty left by a history from a learner's ability to use that
history. A small environment with known outcome probabilities makes
both quantities computable.

\paragraph{Environment and prediction target.}
There are four contexts, two actions, and two equally likely deployments.
In deployment zero, the action-success probabilities are
\[
P_0=\begin{pmatrix}.90&.55\\.75&.40\\.45&.85\\.60&.90\end{pmatrix}.
\]
Deployment one reverses the action columns:
$P_1(x,a)=P_0(x,1-a)$. Contexts and historical actions are sampled
uniformly and independently, and outcomes are independent Bernoulli
draws conditional on the context, action, and deployment.
A historical record is $(x_i,a_i,o_i)$; LimiX-2 receives two context bits
and an action bit and predicts the binary outcome.
It selects
\begin{equation}
\widehat a_D(x)=\arg\max_{a\in\{0,1\}}q_{D,\mathcal T}(O=1\mid x,a).
\label{eq:decision}
\end{equation}
The inputs are numeric, so this example uses LimiX-2 directly.

\paragraph{Reference rules and gaps.}
The oracle observes the active deployment and chooses its best action.
A family-aware Bayes rule knows $P_0$, $P_1$, and the equal prior, and
infers the deployment from the supplied history $D$.
For each rule, risk is expected failure probability under the same
distribution of deployments, contexts, and outcomes. The information
gap is Bayes risk minus oracle risk; the inference gap is LimiX-2 risk
minus Bayes risk. Both use posterior expected risk conditional on $D$,
as in Equation~\ref{eq:gap_decomposition}.
The oracle sees the deployment identity but not future outcome draws.

\paragraph{Sampling and numerical illustration.}
Histories contain 0, 4, 16, 64, or 256 records, with ten sampling seeds
paired across the two deployments. The zero-history condition uses a
fixed-action reference; nonempty histories use LimiX-2 with 32-pipeline
aggregation (Appendix~\ref{app:implementation}).
Success is the known success probability of the selected action,
averaged over contexts. At 256 records, the matching deployment history
gives 85.00\% expected success, compared with 50.00\% for the opposite
deployment's history and 67.50\% for the fixed-action reference.
The supplied history therefore determines which actions are appropriate
for the current deployment.

At 64 records, the mean information and inference gaps are 0.00115
and 2.624 additional failures per 100 decisions, respectively
(95\% paired-seed interval for the inference gap: [1.250, 3.937]).
The interval uses 20,000 bootstrap resamples of the ten seeds,
averaging the two deployments within each seed.
At this history size, the inference gap accounts for most of the excess risk.
At 256 records, LimiX-2 and Bayes choose the same actions in all twenty
deployment--history conditions, and the estimated inference gap is zero.